\documentclass[letterpaper]{article}
\usepackage{aaai}
\usepackage[utf8]{inputenc}
\usepackage{color}
\usepackage{amsmath}
\usepackage{amssymb} 
\usepackage{graphicx,subcaption}
\usepackage[ruled,linesnumbered]{algorithm2e}
\usepackage{wrapfig}

\title{\LARGE \bf Towards Learning Efficient Maneuver Sets for
  Kinodynamic Motion Planning}
\author{Aravind Sivaramakrishnan, Zakary Littlefield and Kostas E. Bekris
\thanks{The authors are with the Department of Computer Science, Rutgers University, Piscataway, New Jersey, 08854, USA. Email: {\tt\small \{as2578,zwl2,kb572\}@rutgers.edu}}%
}
\date{March 2019}

\begin{document}
\newcommand{\argmax}[1]{\underset{#1}{\operatorname{arg}\,\operatorname{max}}\;}
\newcommand{\argmin}[2]{\underset{#2}{\operatorname{arg}\,\operatorname{min}}\;}

\maketitle

\section{Introduction}

Planning for systems with dynamics is challenging as often there is no local planner available and the only primitive to explore the state space is forward propagation of controls. In this context, tree sampling-based planners \cite{LaValle2001,Karaman:2011aa,Webb2013Kinodynamic-RRT} have been developed, some of which achieve asymptotic optimality by propagating random controls during each iteration \cite{Li:2016}. While desirable for the analysis, random controls result in slow convergence to high quality trajectories in practice.

This short position statement first argues that if a kinodynamic planner has access to local maneuvers that appropriately balance an exploitation-exploration trade-off, the planner's per iteration performance is significantly improved. Exploitative maneuvers drive the system to the goal as fast as possible given local obstacle and heuristic information. Exploration maneuvers allow the system to move in a variety of different directions so as to deal with situations that the heuristic does not provide good guidance.  Generating such maneuvers during  planning can be achieved by curating a large sample of random controls. This is, however, computationally very expensive. If such maneuvers can be generated fast, the planner's performance will also improve as a function of computation time.

Towards objective, this short position statement argues for the integration of modern machine learning frameworks with state-of-the-art, informed and asymptotically optimal kinodynamic planners \cite{littlefield2018dirt}. The proposed approach involves using using neural networks to infer local maneuvers for a robotic system with dynamics, which properly balance the above exploitation-exploration trade-off. In particular, a neural network architecture is proposed, which is trained to reflect the choices of an online curation process, given local obstacle and heuristic information. The planner uses these maneuvers to efficiently explore the underlying state space, while still maintaining desirable properties. Preliminary indications in simulated environments and systems are promising but also point to certain challenges that motivate further research in this direction.

\section{Problem Setup}

Consider a robot with a state space $\mathbb{X}$ - divided into a collision-free $\mathbb{X}_f$ and obstacle subset $\mathbb{X}_{obs}$ - and a control space $\mathbb{U}$, which respects dynamics of the form: $\dot{x}=f(x,u)$, where $x \in \mathbb{X}$ and $u \in \mathbb{U}$. The process $f$ can be an analytical ordinary differential equation or modeled via a physics engine.

We will refer to a finite set of piecewise-constant controls $u$, each of them propagated for a specified time duration $t$ as a \textit{maneuver set}, or $U$. When a sequence of maneuvers are executed one after the other, they define a \textit{plan}. A plan of length $T$ induces a trajectory $\pi \in \Pi$ where $\pi: [0,T] \rightarrow \mathbb{X}_f$. 

Each trajectory has a cost $\texttt{C}$ according to function cost : $\Pi \rightarrow \mathbb{R}^+$. The solution trajectory must minimize this cost. A useful technique that aids in finding solutions very quickly is the heuristic function, $h(x)$, which provides guidance towards the goal set $\mathbb{X}_G$. This function is assumed to be an admissible and consistent estimate of the cost to go from every state in $\mathbb{X}_f$ to the goal. 

For kinodynamic planning problems, it has been shown that propagating random controls at every state and appropriately selecting which state is expanded can guarantee asymptotic optimality \cite{Li:2016,hauser2016asymptotically}, i.e., the cost $\texttt{C}$ of the generated trajectory $\pi$ approaches the optimum cost $\texttt{C}^*$ with probability 1 as the number of iterations of the algorithm increase. 

\section{Motion planning with informed maneuvers}

At a high-level, a sampling-based planner, such as RRT \cite{LaValle2001}, builds a tree structure of states reachable from the start and follows a selection-propagation process to expand the tree until it intersects $\mathbb{X}_f$. During each iteration, a tree node is selected and a control is propagated from this state. Certain instances of this framework, such as SST \cite{Li:2016}, identify the conditions for asymptotic optimality (AO), which involves propagating random controls. More recent variants, such as DIRT \cite{littlefield2018dirt}, aim to computational efficiency by using a heuristic, while maintaining the AO property. The heuristic can be used to: a) promote selecting nodes closer to the goal, b) prune nodes that cannot provide a better solution than one already discovered and c) prioritize the propagation of maneuvers among random candidates, which bring the system closer to the goal. 

\begin{figure}[h]
\centering
\includegraphics[width=1.5in]{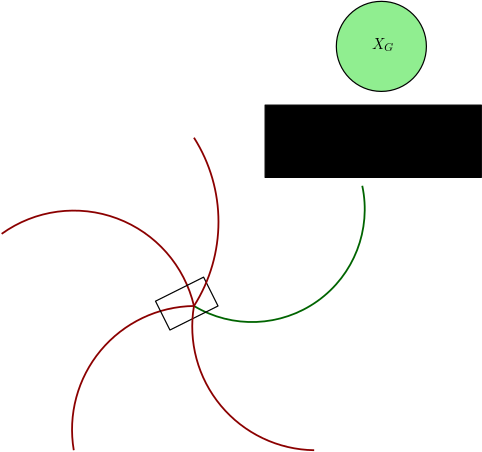}
\caption{Maneuvers for a robot planning to reach $X_G$ (green circle) behind an obstacle (black box). The exploitative control (green) greedily takes the system towards $X_G$. The explorative controls (red) attempt to provide coverage. \vspace{-.2in}}
\label{fig:mset}
\end{figure}

One way that the DIRT algorithm allows to use the heuristic, while maintaining AO, is that it employs a set of informed maneuvers at each state, which are propagated first when the corresponding node is selected, and then it reverts to propagating random controls from the corresponding state. A key question in this context is how to generate the set of informed maneuvers so as to improve the efficiency of the planner relative to using just random controls. A promising way is to compute maneuver sets at each state is to balance exploitation and exploration. These maneuver sets can either be computed online, as we do in this work, or pre-computed motion primitives can be used. \cite{pivtoraiko2011kinodynamic} On one hand, exploitation maneuvers should guide the system as much as possible towards the goal given local heuristic information. On the other hand, exploration maneuvers should drive the system in a diverse set of states to discover alternative paths when the heuristic guidance is incorrect. 

This section describes a naive and computationally expensive process for generating such maneuvers during the online operation of the planner. In particular, a sufficiently large set of random controls are first propagated from the selected state. This set of random controls is then curated to form a \textit{curated maneuver set}, as outlined in Fig~\ref{fig:mset}.

In more detail, a set of trajectories are generated by randomly sampling control sequences $U_{cand}=\{\upsilon_0,\upsilon_1,\cdots,\upsilon_{M}\}$ and forward simulating each control for a specified duration from a given state to generate a trajectory set $\Pi_{cand}=\{\pi_0,\pi_1,\cdots,\pi_{M}\}$.  To determine the \textit{exploitative trajectory} from the set $\Pi_{cand}$, the one that gets closest to the goal state without colliding with any obstacles is chosen:

\vspace{-.1in}
$$\argmin_{\pi_{new} \in \Pi_{cand}} h(\pi_{new}(T))$$
\vspace{-.05in}

The control that generated this exploitative trajectory, $u^0$ is added to the curated maneuver set $\hat{U}$. The remaining $N-1$ \textit{exploratory} controls $u^1,\cdots,u^N$ for the maneuver set $\hat{U}$ are added incrementally. The set of trajectories that result from a given start state $x$ using maneuver set $\hat{U}$ is denoted by $\hat{\Pi}$. Among all the trajectories in $\Pi_{cand}$, the trajectory that maximizes a trajectory dispersion metric $d_\pi({\cdot,\cdot})$ from the previously selected trajectories in $\hat{\Pi}$ is added to $\hat{\Pi}$:

\vspace{-.1in}
$$\argmax{\pi_{new}\in\Pi_{cand}}\min_{\pi\in\hat{\Pi}}\ d_\pi(\pi_{new},\pi).$$
\vspace{-.05in}


For generating the exploratory maneuvers, this work proposes to employ a metric similar to the one considered by Green and Kelly \cite{green2007toward}. This previous approach uses a distance metric $d_\pi(\pi_1,\pi_2)$ to determine how similar two trajectories are to one another. The distance metric between trajectories considered here is an approximation of the area between two trajectories. Using this metric, it is possible to find trajectories that are sufficiently dispersed from one another to create a wide spanning set of trajectories. While the prior work  \cite{green2007toward} aimed to create a single set of trajectories offline that would solve a wide variety of problems, the approach here is used to generate a set of controls that is tailored to each state of the robot during planning that is selected for propagation. This strategy provides good adaptability to the environment and is appropriate for robots with high order dynamics where a single offline maneuver set may not be expressive enough to solve most motion planning queries.
					
\vspace{-0.1in}
\begin{table}[h!]
\centering
\begin{tabular}{|l|l|l|l|}
\hline
        & \textbf{Iteration} & \textbf{Comp. Time} & \textbf{Path Cost} \\ [0.5ex] \hline
Random  & 1471                    & 0.2                   & 50.47                  \\ \hline
Curated & 686                    & 12.15                & 48.13                  \\ \hline
\end{tabular}
\caption{Statistics for finding the first solution path between DIRT using random and curated maneuvers.}
\label{table:Rastar}
\end{table}
\vspace{-0.1in}

Table~\ref{table:Rastar} studies the behavior of the DIRT planner with randomly generated and curated maneuvers. Though the curated procedure is very effective in finding a high-quality solution and can do so very fast in terms of a per iteration performance, the computational cost of generating the curated maneuver set online quickly becomes prohibitive. Hence, it is desirable to develop an approach that achieves the same objective as the curation but can generate the maneuvers fast. The curated planner can be used to generate a dataset of curated maneuvers for a robot in a simulated environment. Then, data-driven machine learning techniques can be employed to learn how to generate maneuver sets online at a similar computational cost to selecting random controls.

\begin{figure}[h!]
	\centering
	\begin{subfigure}{0.4\columnwidth} 
		\centering
		\includegraphics[width=\linewidth]{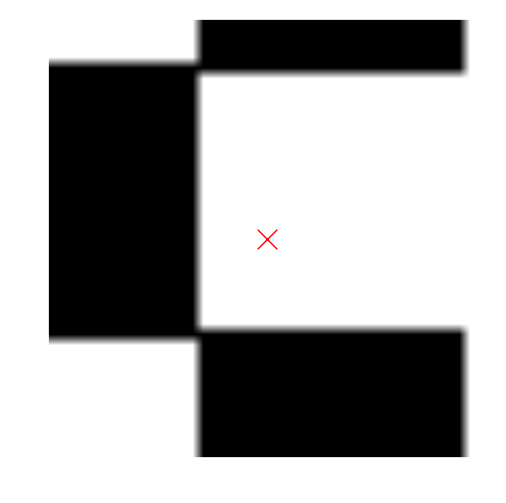}
		\caption{Example $o_{local}$. The light region represents $X_{f}$, while the dark region represents $X_{obs}$.} 
	\end{subfigure}
	\hfill
	\begin{subfigure}{0.4\columnwidth} 
		\centering
		\includegraphics[width=\linewidth]{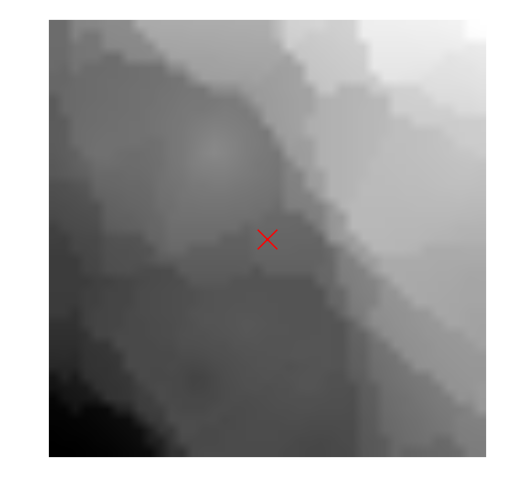}
		\caption{Example $h_{local}$. The lighter regions are regions of decreasing heuristic magnitude.} 
	\end{subfigure}
	\caption{Examples of $o_{local}$ and $h_{local}$. In both cases, the project of the robot state $x_c$ in the workspace has been marked with an \textcolor{red}{$\times$}. \vspace{-.2in}} 
	\label{fig:ex_data}
\end{figure}

\begin{figure*}[h!]
\centering
\includegraphics[scale=0.4]{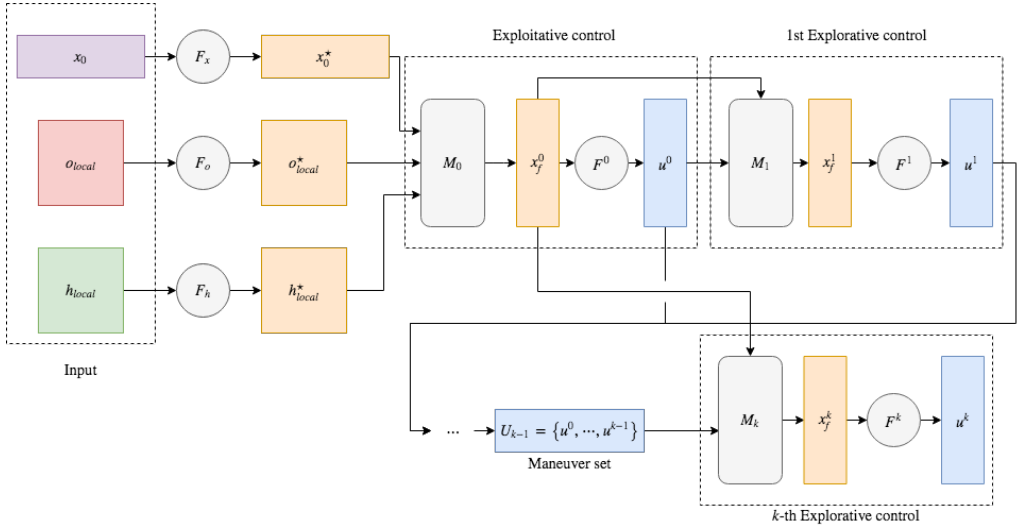}
\vspace{-.1in}
\caption{Computation graph of $U_k = \hat{f}(x_0,o_{local},h_{local})$. For $k=N, U_k = \hat{U} = \{u^0,\cdots,u^N\}$.\vspace{-.15in}}
\label{fig:cgraph}
\end{figure*}

\section{Learning promising maneuvers}

The new objective is to learn a maneuver set $\hat{U} = \{u^0, u^1,\cdots,u^N\}$, where $u^0$ is the exploitative control, and $u^1,\cdots,u^N$ are the exploratory ones. Any control $u^k \in U$ must aid the planner in finding a collision-free trajectory $\pi$ from the currently selected state $x_c$ to the desired goal $x_G$. 

The input to the learning process is constructed as follows. First, a regular set of points $\mathbb{X}_{local}$ in the vicinity of $x_c$ are collision checked to generate a binary 2D image $o_{local}(x_c)$ indicating the presence of obstacles in the workspace (currently the focus is on navigation challenges, where the workspace projection has a major impact on the collision properties of robot states). The heuristic $h(x)$ is also evaluated at each $x \in \mathbb{X}_{local}$, resulting in a 2D matrix $h_{local}(x_c)$. Examples of  $o_{local}$ and $h_{local}$ are shown in Fig~\ref{fig:ex_data}.

In addition to $x_c$, providing $h_{local}$ as input to the model helps bias the model to predict an exploitative control $u^0$ that takes the system closer to the goal. The $o_{local}$ map is used as input to discourage the model from generating controls that may steer the robot in directions where obstacles are present. 


Figure \ref{fig:cgraph} highlights the proposed architecture. The non-convex functions $F_x, F_o, F_h$, represented by multi-layered neural networks, act on the initial state $x_c$ (or $x_0$ in the image), the local obstacle information $o_{local}$, and the local heuristic gradient $h_{local}$ respectively, to produce the high-level feature representations $x_0^*, o_{local}^*$ and $h_{local}^*$. An operator $M_0(x_0^*, o_{local}^*,h_{local}^*)$ then produces a feature vector $x_f^0$, from which the first exploitative control $u^0$ is obtained as $u^0 = F^0(x_f^0)$, where $F^0$ is also a non-convex function represented as a multi-layered neural network. 

Once the first exploitative control is obtained, the remaining $N$ exploratory controls are obtained as follows.

\vspace{-.1in}
\begin{align}
	x_f^k &= M_k(x_f^0,U_{k-1}) \\
	u^{k} &= F^k(x_f^k) 
\end{align}
\vspace{-.1in}

\noindent where for all $k \geq 1$, $U_k = \{u^0,u^1,..,u^{k-1}\}$, $M_k$ is an operator that acts on $x_f^0$ and $U_{k-1}$ to produce a feature vector $x_f^k$, and $u^k$ is obtained as above. In this work, $F^k$ is a multi-layered neural network. For the exploitative control ($k=0$), $U_{k-1}$ is the empty set.

In the experiments, the functions $F^0,\cdots,F^N$ are all multi-layered perceptrons with one hidden layer activated by the Rectified Linear Unit (ReLU) activation function and one output layer. The operator $M_k$ is chosen to be a simple concatenation operator. The operators $M_1,\cdots,M_N$ drop the learned heuristic representation $h_{local}^*$ as it is not useful in predicting explorative controls. Two types of networks are considered in the experiments: \texttt{FC} and \texttt{Conv}, which use fully connected layers and convolutional layers respectively for the functions $F_o$ and $F_h$.

\section{Evaluation}

The evaluation considers a treaded vehicle with dynamics from the literature \cite{pentzer2014model}. The state space is 5 dim., corresponding to  SE(2) augmented by the steering angle and the forward velocity state variables. The controls are 2 dim. and correspond to the accelerations on the left and right treads.

\begin{table*}[]
\centering
\begin{tabular}{|l|l|l|l|l|l|}
\hline
\textbf{Algorithm}    & \textbf{NumSolns} & \textbf{FirstSolnIters} & \textbf{FirstSolnCost} & \textbf{FinalSolnIters} & \textbf{FinalSolnCost} \\ [0.5ex]  \hline
DIRT - Random         & 30                    & 3446.67                 & 59.64                     & 23277.57                & 49.44                     \\ \hline
DIRT - FC (Exploit)   & 30                    & 2246.67                 & 56.54                     & 17050.37                & 49.89                     \\ \hline
DIRT - FC (All)       & 30                    & \textbf{620}                     & \textbf{47.58}                     & \textbf{16921.5}                 & \textbf{45.47}                     \\ \hline
DIRT - Conv (Exploit) & 30                    & 3366.67                 & 65.03                     & 27774.67                & 48.38                     \\ \hline
DIRT - Conv (All)     & 30                    & 2006.67                 & 54.8                      & 25671.07                & 48.16                     \\   \hline
\end{tabular}
\vspace{-.1in}
\caption{Solution statistics for \texttt{Greedy}. All values are averaged over \texttt{NumSolns}. Best values highlighted in bold.
\vspace{-.1in}}
\label{table:iai}
\end{table*}

\begin{table*}[]
\centering
\begin{tabular}{|l|l|l|l|l|l|}
\hline
\textbf{Algorithm}    & \textbf{NumSolns} & \textbf{FirstSolnIters} & \textbf{FirstSolnCost} & \textbf{FinalSolnIters} & \textbf{FinalSolnCost} \\
[0.5ex] \hline
DIRT - Random         & 30                & 15666.67                & 163.60                 & 33254.13                & 149.47                 \\ \hline
DIRT - FC (Exploit)   & 29                & \textbf{12000}                   & 155                    & 31794.86                & 140.06                 \\ \hline
DIRT - FC (All)       & 30                & 18766.67                & \textbf{133.83}                 & \textbf{28119.66}                & \textbf{130.92}                 \\ \hline
DIRT - Conv (Exploit) & 29                & 27666.67                & 182.16                 & 39924.96                & 172.14                 \\ \hline
DIRT - Conv (All)     & 30                & 14066.67                & 143.71                 & 28194.83                & 139.43                 \\ \hline
\end{tabular}
\vspace{-.1in}
\caption{Solution statistics for \texttt{Explore}. All values are averaged over \texttt{NumSolns}. Best values highlighted in bold.
\vspace{-.1in}}
\label{table:explore}
\end{table*}

\begin{figure}[h!]
\vspace{-.1in}
	\centering
	\begin{subfigure}{0.4\columnwidth} 
		\centering
		\includegraphics[width=\linewidth]{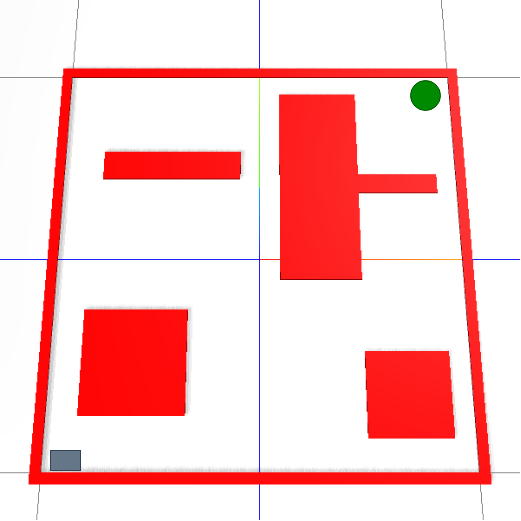}
		\caption{\texttt{Greedy}} 
	\end{subfigure}
	\hfill
	\begin{subfigure}{0.4\columnwidth} 
		\centering
		\includegraphics[width=\linewidth]{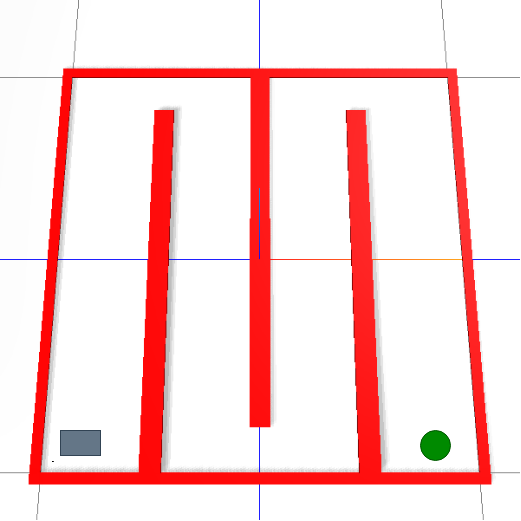}
		\caption{\texttt{Explore}} 
	\end{subfigure}
\vspace{-.1in}
	\caption{Environments: The grey rectangle us the starting pose of the robot (facing right) and the green circle is the goal region. The robot must avoid the red obstacles.
\vspace{-.15in}} 
	\label{fig:envs}
\end{figure}

For training, obstacles are randomly placed in a 2 dim. workspace so they cover one-third of the reachable workspace. The DIRT planner \cite{littlefield2018dirt} is executed with the online curation procedure on multiple problem instances in such workspaces. The Euclidean distance between two points in the workspace is used as the heuristic function, and the duration of the solution trajectory as the cost function. For each node $x_c$ the planner selects to propagate, the training process stores the  $o_{local}$ and $h_{local}$ maps, as well as a maneuver set $\hat{U}$ of size 5 (1 exploitative control and $N = 4$ exploratory controls). This maneuver set is curated from a set of 1000 randomly sampled maneuvers. Two environments are considered: a simple environment where the heuristic guides the robot effectively to the goal (\texttt{Greedy}), and an adversarial environment where acting greedily does not enable the robot to reach the goal (\texttt{Explore}). These environments are illustrated in Fig~\ref{fig:envs}.

Tables~\ref{table:iai} and ~\ref{table:explore} compare the performance of DIRT (after 50k iterations) for the following maneuvers: a) random (DIRT - Random), b) exploitative control predicted either by a Fully Connected (DIRT - FC (Exploit)) or a Convolutional network (DIRT - Conv (Exploit)), and c) both exploitative and explorative controls are predicted by the networks ((DIRT - FC (All), DIRT - Conv (All)). The following metrics are measured: (a) success rate over 30 experiments (NumSolns), (b) in how many iterations does the planner find a solution (FirstSolnIters) and of what quality (FirstSolnCost), and (c) similar statistics for the final solution (FinalSolnIters, FinalSolnCost). 

DIRT - FC (All) obtains the highest quality solution. The number of iterations required for the best solution is also the lowest for DIRT - FC (All), while the iterations required for the first solution is lower for DIRT - FC (Exploit) in the \texttt{Explore} environment and DIRT - FC (All) in the \texttt{Greedy} environment. Using only the exploitative control affects the online performance of the planner in the \texttt{Explore} environment, where it is not able to find a solution in every run. 

\section{Discussion}
There are indications that the learned maneuvers guide the vehicle to the goal effectively but frequently result in collisions, which partly degrades performance. An improved learning process is necessary to increase the rate of collision-free maneuvers. Furthermore, the current cost of network inference is more expensive than returning a random control. This makes the fully connected network return the first path in similar time to random maneuvers, while the convolutional network takes significantly longer. No special purpose hardware, however, was used for the current experiments, such as GPUs, which can speed up network inference.  The effectiveness of the exploration controls could be further improved by considering not only the controls propagated at the current state but also the tree nodes in the vicinity of the propagation. As the proposed pipeline is scaled to higher dimensional systems, more complex environments and realistic sensing input, there are additional considerations related to data efficiency and uncertainty that must be mitigated.

\bibliographystyle{aaai}

\end{document}